\documentclass[letterpaper]{article} 
\usepackage{aaai25}  
\usepackage{times}  
\usepackage{helvet}  
\usepackage{courier}  
\usepackage[hyphens]{url}  
\usepackage{graphicx} 
\usepackage{natbib}  
\usepackage{caption} 
\usepackage{algorithm}
\usepackage{algorithmic}
\usepackage{multirow}
\usepackage{booktabs}
\usepackage{amsmath} 
\usepackage{amsfonts}
\usepackage{amssymb}
\usepackage{svg}
\usepackage{enumitem}
\usepackage{subcaption}
\usepackage{newfloat}
\usepackage{listings}
\DeclareCaptionStyle{ruled}{labelfont=normalfont,labelsep=colon,strut=off} 
\floatstyle{ruled}
\newfloat{listing}{tb}{lst}{}
\floatname{listing}{Listing}
\title{Multi-Task Curriculum Graph Contrastive Learning \\ with Clustering Entropy Guidance}
\author{
	Chusheng Zeng\textsuperscript{\rm 1}, Bocheng Wang\textsuperscript{\rm 1}, Jinghui Yuan\textsuperscript{\rm 1,\rm 2}, Rong Wang\textsuperscript{\rm 1}, Mulin Chen\textsuperscript{\rm 1}\thanks{Corresponding author}
}
\affiliations{
    \textsuperscript{\rm 1}School of Artificial Intelligence, OPtics and ElectroNics (iOPEN),\\
 Northwestern Polytechnical University, Xi’an 710072, Shannxi, China\\
    \textsuperscript{\rm 2}Institute of Artificial Intelligence (TeleAI), China Telecom, P. R. China \\
    zcs@mail.nwpu.edu.cn, wangbocheng@mail.nwpu.edu.cn, yuanjh@mail.nwpu.edu.cn\\ wangrong07@tsinghua.org.cn, chenmulin001@gmail.com

}

\usepackage{bibentry}

\begin{document}

\maketitle

\begin{abstract}
Recent advances in unsupervised deep graph clustering have been significantly promoted by contrastive learning. Despite the strides, most graph contrastive learning models face challenges: 1) graph augmentation is used to improve learning diversity, but commonly used random augmentation methods may destroy inherent semantics and cause noise; 2) the fixed positive and negative sample selection strategy is limited to deal with complex real data, thereby impeding the model's capability to capture fine-grained patterns and relationships. To reduce these problems, we propose the Clustering-guided Curriculum Graph contrastive Learning (CCGL) framework. CCGL uses clustering entropy as the guidance of the following graph augmentation and contrastive learning. Specifically, according to the clustering entropy, the intra-class edges and important features are emphasized in augmentation. Then, a multi-task curriculum learning scheme is proposed, which employs the clustering guidance to shift the focus from the discrimination task to the clustering task. In this way, the sample selection strategy of contrastive learning can be adjusted adaptively from early to late stage, which enhances the model's flexibility for complex data structure. Experimental results demonstrate that CCGL has achieved excellent performance compared to state-of-the-art competitors. 
\end{abstract}

%

\section{Introduction}
In recent years, the representational capacity of graph data has made it prevalent in various applications, including social networks, knowledge graphs, and traffic prediction. The prevalence has been further amplified by the emergence of deep graph neural networks \cite{kipf2016semi,velivckovic2017graph}, which have enabled the efficient analysis of complex graph data. Specifically, unsupervised deep graph learning \cite{wang2019attributed,liu2022towards,mo2022simple} has garnered extensive research interest due to its ability to extract  discriminative, and interpretable graph features.

Graph contrastive learning is a fundamental paradigm within the field of unsupervised graph learning, and consists of two parts including graph augmentation and contrastive learning. \textbf{Data augmentation} changes the original data graph to obtain multiple views with similar semantics, thereby expanding the selection space of positive and negative sample pairs. For example, many data augmentation methods select the nodes and edges randomly, and delete$/$mask$/$disturb them to produce an augmented graph \cite{zhu2020deep, you2020graph}. \textbf{Contrastive learning} enhances the similarity of samples with related semantics, while pushing samples with low  relevance away from each other. The above process helps to learn graph embeddings with high discriminability, which in turn reveals the implicit structure of the data. The early methods \cite{DBLP:conf/iclr/HjelmFLGBTB19, veličković2018deepgraphinfomax} take the non-corresponding nodes among views as negatives, which can mistakenly pull many intra-class samples far apart, resulting in sampling bias. Some recent methods \cite{lin2022prototypical,zhao2021graph} have attempted to improve the selection strategies for positive and negative samples, adopting specific strategies to select more appropriate positive and negative samples, such as taking the intra-class nodes as positive and the inter-class nodes as negative.

Existing methods for graph contrastive learning often encounter difficulties when dealing with complex data. On the one hand, the widely used random graph data augmentation methods \cite{rong2019dropedge, feng2020graph,fang2023dropmessage} usually corrupt the fundamental characteristics of the data and may lead to noise, while other data augmentation methods mainly rely on artificial priors and may be unsuitable for the clustering task. On the other hand, most methods \cite{you2021graph,chen2024deep} adopt the fixed strategy to select positive and negative samples throughout the training process, which is not flexible in practical applications. For example, some methods\cite{yang2023cluster, lin2022prototypical,xia2022self} select samples based on pseudo labels, which may be unreliable in the early stages due to the insufficient discriminability of the learned embeddings. The limitation hinders the model's ability to explore the intrinsic representation and complex topological relationship.

To alleviate the above problems, we propose Clustering-guided Curriculum Graph contrastive Learning (CCGL) framework. As shown in Fig. 1, the intermediate results of the model are clustered to calculate the clustering entropy, which serves as a clustering guidance for the whole framework. On this basis, a clustering-friendly strategy is adopted for performing structure-level and feature-level data augmentation. Then, the multi-task curriculum learning scheme uses clustering entropy to determine the clustering confidence of samples and place them in different contrastive learning tasks. As training progresses, samples are transited from the early stage simple discrimination task to the more challenging clustering task in the late stages, such that the flexible adjustment of sample selection is achieved. The main contributions of this paper are as follows.

\begin{itemize}
\item A clustering-guided curriculum graph contrastive learning framework is established. The clustering entropy is defined to serve as the clustering guidance, which is used to evaluate the importance and clustering confidence of nodes throughout the whole framework.
\item A clustering-friendly graph augmentation method is proposed. Under the clustering guidance, structure augmentation tends to preserve intra-class edges, while feature augmentation is more likely to retain the class-specific features.
\item A multi-task curriculum learning scheme is developed to explore the complex data structure. It allows the model to first learn discriminative representations of the samples and then move towards clustering, enhancing the capability to capture cluster-oriented discriminative features.
\end{itemize}

\section{Related Work}

\subsection{Graph Contrastive Learning}

After making progress in image recognition\cite{chen2020simple, zbontar2021barlow, wang2022contrastive}, contrastive learning combined with Graph Neural Networks (GNNs) has also garnered significant attention from many researchers. Explorations in graph contrastive learning have primarily focused on graph augmentation and contrastive objectives. 

Data augmentation enriches the diversity of training samples by generating different views. MVGRL \cite{hassani2020contrastive} and DCRN \cite{liu2022deep} utilize graph diffusion to create augmented views, while methods like GRACE \cite{zhu2020deep} and SCAGC \cite{xia2022self} achieve this through random attribute and edge perturbations. Most augmentation techniques are often stochastic and uncontrollable, potentially disrupting semantics and cause noise. Recently, several adaptive augmentation methods have been proposed. GCA \cite{zhu2021graph} learns the weights of discarding adaptively, with the expectation that the model's final learned representations will be insensitive to unimportant nodes or edges. However, It is not necessarily suitable for downstream clustering tasks. CCGL is designed with clustering as the objective, aiming to obtain augmented views that better align with the underlying clustering structure.

Contrastive objective learns the embedding by constructing pairs of positive and negative samples. MVGRL designs an InfoMax loss to maximize the cross-view mutual information between nodes and the global summary of the graph. AGE\cite{cui2020adaptive} devises a pretext task, using a cross-entropy loss to classify similar and dissimilar nodes. Based on $K$-Means or other graph-based clustering methods\cite{yuan2024doubly}, some methods\cite{lin2022prototypical,yang2023cluster} use intermediate clustering results to guide the contrast objective. CCGL designed two types of contrastive objectives to tackle tasks of varying difficulties, and we employ curriculum learning to tailor the tasks for each node based on the model's current learning state.

\subsection{Curriculum Learning}
Curriculum learning \cite{bengio2009curriculum, wang2021survey} is a training strategy that mimics the human learning process. It allows the model to start with simpler samples and gradually progress to more complex ones, as well as to advanced knowledge. 
Some studies \cite{jiang2018mentornet, han2018co} have theoretically demonstrated the effectiveness of curriculum learning in enhancing generalization capabilities when dealing with noisy data. Curriculum learning is widely applied across machine learning\cite{zbontar2021barlow} and deep learning\cite{matiisen2019teacher,graves2017automated}. 

Curriculum Learning consists of two main parts: difficulty measurer and training scheduler\cite{hacohen2019power}. The difficulty measurer is used to evaluate the difficulty of the sample. Predefined difficulty measurer \cite{platanios2019competence,spitkovsky2010baby,tay2019simple} is mainly designed manually according to the data characteristics of a specific task. The training scheduler determines the appropriate training data to feed into the model based on the evaluation results of the difficulty measurer. But existing predefined training schedulers\cite{cirik2016visualizing} are usually data and task independent, and most curriculum learning in various scenarios uses a similar training scheduler. The discrete scheduler adjusts the training data at each fixed round or when the current data converges, while the continuous scheduler adjusts the training data at each round according to a defined scheduling function. Different from the general curriculum learning that only controls the number of sample participation, our proposed multi-task curriculum learning scheme lets samples perform tasks of different difficulty at different stages of the model, which considers the difficulty distribution of samples, leading to improved generalization capabilities.

\section{Methodology}
In this section, we will provide a detailed introduction to the proposed CCGL method.

\begin{figure*}[h] 
\centering
\includegraphics[width=0.9\textwidth]{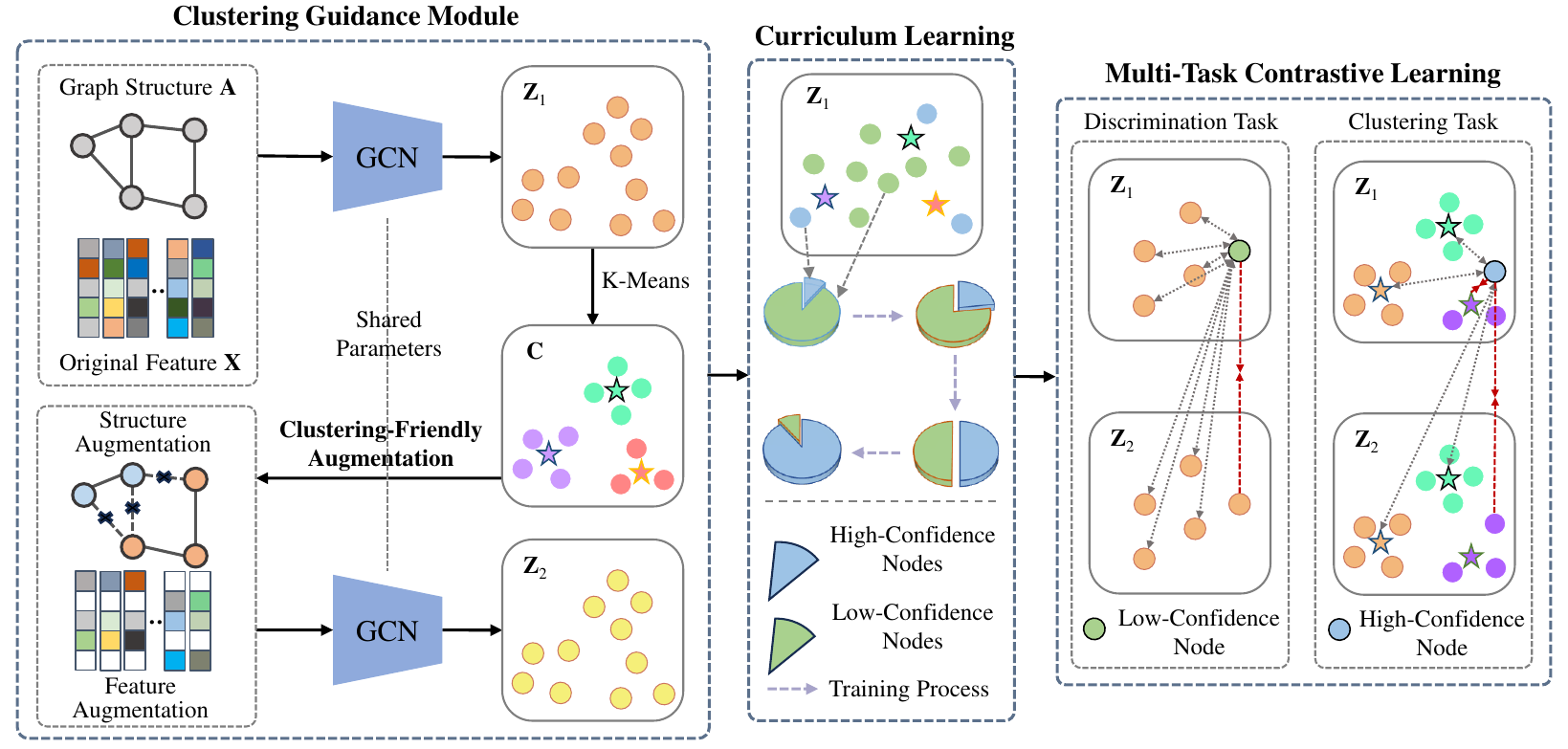}
\caption{The pipeline of CCGL. The Clustering Guidance Module clusters the embedding $\mathbf{Z_{1}}$ to obtain clustering guidance. Clustering-Friendly
Augmentation applies clustering-oriented structure augmentation and feature augmentation to the original data. According to the clustering guidance, Crriculum Learning divides the nodes into high confidence groups and low confidence groups to perform different contrastive tasks in Multi-Task Contrastive Learning.}
\label{fig:my_svg_image}
\end{figure*}

\subsection{Notations and Problem Definition}
In this paper, uppercase letters represent matrices and lowercase letters represent vectors. An undirected graph with
$n$ nodes is defined as $\mathcal{G} = \{{\mathbf{X}}, {\mathbf{A}}\}$, and $\mathbf{X} \in \mathbb{R}^{n \times d}$ is the original feature matrix of the nodes, ${\mathbf{A}} \in \mathbb{R}^{n \times n}$ represents the adjacency matrix of the graph data.


This paper aims to address an unsupervised graph clustering problem. Specifically, given a graph $\mathcal{G} = \{{\mathbf{X}}, {\mathbf{A}}\}$, the task is to train a GCN (Graph Convolutional Neural Networks) encoder such that ${\mathbf{Z}} = f({\mathbf{X}},{\mathbf{A}})$, and the resulting ${{\mathbf{Z}}}$ can be used to cluster all nodes $\mathcal{V}$ of the graph $\mathcal{G}$ into $k$ classes.

\subsection{Clustering Guidance Module}
The clustering guidance module is introduced to make the learned embeddings more suitable for clustering tasks. Clustering entropy is defined to guide subsequent augmentation and contrastive learning. It also serves as a loss function to improve the quality of pseudo-labels.

Given the original graph data $\mathcal{G} = \{{\mathbf{X}}, {\mathbf{A}}\}$ and embedding $\mathbf{Z_{1}}$ computed by GCN, performing $k$-Means on $\mathbf{Z_{1}}$ results in $k$ cluster centroids \( {\mathbf{C}} = \{ \{ {c}_1 \}, \{ {c}_2 \}, \ldots, \{ {c}_k \} \} \) and pseudo-labels \( {L} \) of the samples. Then, we calculate the probability assignment matrix \( \mathbf{{P}} \) by
\begin{equation}
\mathbf{P} = {softmax}({\mathbf{Z_{1}}} \cdot \mathbf{{C}^T} ),
\end{equation}
where \( \mathbf{{Z_{1}}\cdot{C}^T} \) calculates the dot product similarity between each sample and the cluster center, the ${softmax}(\cdot)$ performs exponential normalization of each row of \( \mathbf{{Z_{1}}\cdot{C}^T} \) to obtain the probability assignment matrix $ {\mathbf{P}}\in \mathbb{R}^{n \times k}$. Each row of \( \mathbf{{P}} \) represents the probability that the corresponding node is assigned to all the centroids.

To assess the quality of the clustering, we propose the clustering entropy
\begin{equation}
    {E_i} = -\sum_{j=1}^{k} P_{ij} \log(P_{ij}).
\end{equation}
\( {E_i} \) indicates the uncertainty of the probability distribution for each node. Nodes with high \( {E_i} \) have probabilities that are relatively close to each cluster centroid, indicating a lower confidence in the clustering. Conversely, nodes with low \( {E_i} \) have a higher confidence in clustering. For better clustering, the samples with high confidence should increase during the training procedure, such that a low clustering entropy can be achieved. Therefore, a clustering entropy loss is designed to optimize the clustering effect, defined as
\begin{equation}
\mathcal{L}_{EN}=\frac{1}{n}\sum_{i=1}^{n}{E_i}.
\end{equation}
Optimizing $\mathcal{L}_{EN}$ can drive the learned embedding $\mathbf{Z}$ to present a clear cluster structure, and provide more reliable clustering-oriented guidance for the subsequent modules.

\subsection{Clustering-friendly Augmentation}
In this part, we design a clustering-friendly augmentation method, which consists of structure augmentation and feature augmentation. They preserve edges between nodes of the intra-class and important features that are beneficial for the clustering task respectively.

\subsubsection{Structure Augmentation.}
Each edge is assigned with an importance weight to determine its probability of being deleted, which can be formulized as
\begin{equation}
u^{ij}_{e} = {E}_{i}{E}_{j} + \mu \delta ({L}_{i}-{L}_{j}, 0),
\end{equation}
where \( {u}^{ij}_{e} \) represents the importance measure between nodes $i$-th and \( j \)-th. The indicator function \( \delta ({L}_{i}-{L}_{j},0) \) is 0 when \( {L}_{i}={L}_{j} \) and 1 otherwise. The meaning of \( {u}^{ij}_{e} \) is that when the nodes at both ends of the edge have low clustering entropy and belong to the same class, the value of \( {u}^{ij}_{e} \) is relatively smaller, indicating a higher importance of the edge, and thus a lower probability of being removed. After normalization, the probability of each edge being removed is obtained, and the normalization method is given by
\begin{equation}
p_{u}^{e} = \min \left( \frac{u_{e}^{\max }-u_{e}}{u_{e}^{\max }-\bar{u}_{e}} \cdot p_{e}, p_{\tau} \right), 
\end{equation}
where \( u_{e}^{\max} \) and \( \bar{u}_{e} \) are the maximum and average values of \( u_{e} \) respectively. \( p_{e} \) is the overall edge deletion probability, and \( p_{\tau} \) is the truncation probability, which is used to prevent the deletion probability of certain edges from being too high.

The structure augmentation emphasizes the edges that connect reliable nodes of the same class. As a result, important clustering-relevant semantic information is preserved.

\subsubsection{Feature Augmentation.}
Feature augmentation aims to  preserve the representative class-specific features. For one-hot encoded features, 1 indicates the presence of a feature, while 0 indicates its absence, the frequency of each feature appearing in nodes within the same class reflects the importance of that feature. Therefore the feature weight for the $j$-th class ${f}_{j}$ can be defined as
\begin{equation}
{f}_{j} = \sum_{l_i = j} {z}_{i}.
\end{equation}
For features that are not one-hot encoded, the distribution of features within the class can also be used to obtain $\mathbf{f}_{j}$. If the distribution of a feature is more concentrated among the samples within a class, it will have a higher feature weight ${f}_{j}$, which can be determined by statistical measures such as variance.

We can obtain the probability that each feature within each class is masked by normalizing ${f}_{j}$,
\begin{equation}
p_{u}^{f_{j}} = \min \left( \frac{f_{j}^{\max} - f_{j}}{f_{j}^{\max} - \bar{f}_{j}} \cdot {p}_{f}, {p}_{\tau} \right),
\end{equation}
where \( {p}_{f} \) is the overall probability of feature augmentation and \(p_{\tau}\) is the truncation probability.

The augmentation approach emphasizes important features within each cluster, which makes it more convenient to judge the subordinate cluster of each sample. Features irrelevant to clustering are more likely to be removed, thereby reducing noise.

\subsection{Multi-Task Curriculum Learning}
The multi-task curriculum learning scheme is proposed to deal with complex real data. In order to simulate the real-world knowledge learning process, samples should start with simple contrastive task and gradually turn to complex contrastive task. In the early stages of training, the embeddings are not discriminable, and unsuitable for clustering. Therefore, we start from the discrimination task, and end with the clustering task.

\subsubsection{Self-Paced Curriculum Learning.}
In each training iteration of the model, we categorize samples into low-confidence and high-confidence groups to perform the discrimination and clustering task, respectively. A self-paced curriculum learning is used to achieve automatic transformation of tasks according to the clustering entropy.

To distinguish the confidence of samples, we define a indicator vector $v \in \{0,1\}^n$, where \( {v}_{i} = 0 \) denotes that the $i$-th sample belongs to the low-confidence group. Self-paced curriculum learning assigns the contrastive task to samples by controlling each element in the indicator vector $v$. Given $n$ samples, we define $n_{CT}^{t}$ as the number of samples in the model at the $t$-th epoch iteration that are selected to perform the clustering task. The samples are chosen based on their ranking in clustering entropy, and their corresponding elements in the indicator vector $v$ are set to 1. Moreover, $n_{DT}$ is the number of samples that perform discrimination task, and $n_{CT}+n_{DT}=n$. Curriculum pace $\varepsilon$ is defined to control the number of high-confidence samples. With the total number of iterations $T$, the $n_{CT}^{t+1}$ is computed as
\begin{equation}
n_{CT}^{t+1}=\min(n_{CT}^{t}+\varepsilon\frac{n}{T},n).  
\end{equation}
When $\varepsilon$ reaches 1, all samples participate in the clustering task, the $n_{CT}$ will not increase any more.

\subsubsection{Discrimination Task.}
The purpose of discrimination task is to distinguish each sample individually and learn clear self-representations. Since the discrimination task does not consider the topological relationships between samples, it is regarded as a relatively simple task. Samples in the low-confidence group do not exhibit well-defined clustering structures, so we assign them to perform the discrimination task.

To push nodes apart in the embedding space, the positive and negative sampling corresponding to the discrimination task is that the positive sample is the augmented view of the sample itself, while the negative samples are all other nodes. For $z_{i}$, the discrimination task loss is
\begin{align}
&\ell_{DT(i)}=-\frac{1}{2}\sum_{j=1}^{2}\log( \notag \\&
\frac{e^{\mathcal{S}\left(\left(z_{i}^1,z_{i}^2\right)/\tau\right)}}{e^{\mathcal{S}\left(\left(z_{i}^1,z_{i}^2\right)/\tau\right)}+\sum_{k\neq i}e^{\mathcal{S}\left(\left(z_{i}^j,z_{k}^1\right)/\tau\right)}+\sum_{k\neq i}e^{\mathcal{S}\left(\left(z_{i}^j,z_{k}^2\right)/\tau\right)}}),
\end{align}
where \( z_{i}^{j} \) denotes the $i$-th node in the $j$-th view, \( \tau \) is the temperature parameter, and \( \mathcal{S}(\cdot) \) is the similarity calculation function. The overall discrimination task loss can be expressed as
\begin{equation}
\mathcal{L}_{DT}=\frac{1}{n_{DT}}\sum_{i=1}^{n} (1-v_{i})\ell_{DT(i)}. 
\end{equation}
By minimizing $\mathcal{L}_{DT}$, the low-confidence sample gradually captures the key internal features, resulting in a more discriminative embedding. As the discriminability improves, the topological relationship becomes more clear, and the samples gradually transform into high-confidence ones. Then, they can participate into the clustering task.

\subsubsection{Clustering Task.}
Once the samples in the embedding space exhibit sufficient discriminability, the pseudo-labels and -centroids obtained by$k$-Mean become more reliable. Therefore, we can consider the complex topological relationships between samples, aiming to push the intra/inter-class nodes close/disperse for clustering improvement.

For the clustering task, the selection strategy for positive and negative samples is as follows: positive samples are defined as the cluster centroid of the sample and its own augmented view, while negative samples are the centroids of other clusters. For $z_{i}$, the clustering task loss is
\begin{equation}
\mathcal{L}_{CT(i)}=-\frac{1}{2}\sum_{j=1}^2\log( 
\frac{e^{\mathcal{S}\left(\left(z_{i}^1,z_{i}^2\right)/\tau\right)}+e^{\mathcal{S}\left(\left(z_{i}^j,c_{i}\right)/\tau\right)}}{e^{\mathcal{S}\left(\left(z_{i}^1,z_{i}^2\right)/\tau\right)}+\sum_{k}e^{\mathcal{S}\left(\left(z_{i}^j,c_{k}\right)/\tau\right)}}),
\end{equation}
where \( {c}_{i} \) is the pseudo-centroid corresponding to $z_{i}$. Pulling nodes closer to their centroids and pushing them away from other centroids is beneficial to accelerate a clear clustering distribution. The overall loss of clustering contrastive task is
\begin{equation}
\mathcal{L}_{CT}=\frac{1}{n_{CT}}\sum_{i=1}^{n} v_{i}\ell_{DT(i)}.
\end{equation}
By minimizing $\mathcal{L}_{CT}$, the node features with sufficient discrimination after discrimination task learning further show the clustering structure and facilitates clearer clustering separation.

Through the above task transition, the positive and negative sampling strategy gradually changes from considering only the node itself to mining the cluster structure, thus leveraging the clustering information to guide contrastive learning. The progression from easy to challenging tasks enables our method to consistently learn clustering-oriented discriminative features.

\subsection{Joint Loss and Optimization}

Combining Eqs. (3), (9), and (11) , the joint loss is
\begin{equation}
\mathcal{L} = \alpha \mathcal{L}_{DT} + \beta \mathcal{L}_{CT} + \gamma \mathcal{L}_{EN},
\end{equation}
where \( \alpha \), \( \beta \) and \( \gamma \) are hyper-parameters. 

The loss of the model can be considered as a function of the model parameters \( \mathbf{W} \) and the indicator vector \( {v} \), which can be expressed as $\mathcal{L}= g(\mathbf{W}, {v})$. To optimize the objective function, we use an alternate optimization algorithm to iteratively update both. Specifically, $v$ is first initialized as an all-zero vector, which means that all nodes perform the discrimination task at the beginning of the model training. Perform the following two steps alternately until the final iteration. 

Firstly, fix $v^{t}$ and solve $\mathbf{W}^{t+1}$ by
\begin{equation}
\begin{aligned}
& {\mathbf{W}^{t+1}} = \arg \min_{\mathbf{W}^{t}} \big[ \alpha  \mathcal{L}_{CT}(v^{t}, \mathbf{W}^{t}) + \beta \mathcal{L}_{DT}(v^{t}, \mathbf{W}^{t}) \\
& \quad + \gamma \mathcal{L}_{EN}(\mathbf{W}^{t}) \big].
\end{aligned}
\end{equation}
The model parameters $\mathbf{W}^{t+1}$ can be solved using the Adam optimizer. 

Secondly, fix $\mathbf{W}^{t}$ and solve indicator vector $v^{t+1}$ according to the cluster entropy $E$
\begin{equation}
v^{t+1} = \arg \min_{v^{t}} \sum_{i=1}^{N}v_{i}^{t} {E_{i}^{t}}, \text{ s.t. } \| v^{t} \|_{1} = n_{CT}^{t},
\end{equation}
where $n_{CT}^{t}$ initially starts at 0 and increases with the curriculum pace \( \varepsilon \). \( \| v \|_{1} \) is the \( L_1 \) norm of the vector. As \( n_{CT} \) increases, the indicator vector \( v \) will eventually become an all-ones vector, meaning that all nodes will join in the clustering task.

\begin{algorithm}[tb]
\caption{\textbf{CCGL}}
\label{alg:clustering}
\textbf{Input}: Data matrix $\mathbf{X}$, adjacency matrix $\mathbf{A}$, number of classes $k$, curriculum pace $\varepsilon$, edge deletion probability $p_e$, feature masking probability $p_f$.\\
\textbf{Output}: Clustering result.
\begin{algorithmic}[1]
\STATE Initialize model parameters $\mathbf{W}^0$, indicator vector vector $v^0 = 0$, and high-confidence count $n_{CT} = 0$.
\REPEAT
\STATE Compute $\mathbf{Z_{1}} = f(\mathbf{X, A}; \mathbf{W}^t)$.
\STATE Perform $K$-Means on $\mathbf{Z_{1}}$ to obtain clustering entropy $E$ with Eq. (2).
\STATE Apply augmentation to $\mathbf{X,A}$ with the probabilities in Eq. (5) and Eq. (7) to obtain $\mathbf{X', A'}$.
\STATE Compute $\mathbf{Z_{2}} = f(\mathbf{X', A'}; \mathbf{W}^t)$.
\STATE Fix $v^t$, update $\mathbf{W}^t$ with Eq. (14).
\STATE Fix $\mathbf{W}^{t}$, update $v^t$ with Eq. (15).
\STATE Update $n_{CT}$ with Eq. (8).
\UNTIL Reach final iteration $T$.
\STATE \textbf{return} The clustering result in the final round.
\end{algorithmic}
\end{algorithm}

\begin{table}[b]
    \centering
    \setlength{\tabcolsep}{4pt} 
    \begin{tabular}{lcccc} 
    \toprule
    \textbf{Dataset} & \textbf{Samples} & \textbf{Edges} & \textbf{Dimensions} & \textbf{Classes} \\
    \midrule
    CORA & 2708 & 5429 & 1433 & 7  \\
    UAT	& 1190	& 13599 & 239	& 4	\\
    AMAP & 7650	& 119081 & 745 & 8	\\
    AMAC & 13752 & 245861 & 767 & 10 \\
    PUBMED & 19717 & 44438 & 500 & 3 \\
    \bottomrule
    \end{tabular}
    \caption{Descriptions of real-world datasets.}
    \label{table2}
\end{table}

\section{Experiments}



\begin{table*}[h]
    \centering
    \setlength{\tabcolsep}{4pt} 
    \begin{tabular}{l|c|cccccccc|c} 
        \toprule
        {Dataset} &{Metric}  & {K-Means} & {GAE} & {DAEGC} & {SDCN} & {GCA} & {SCAGC} & {CCGC} & {HSAN} & {CCGL}\\
        \toprule
        \multirow{3}{*}{{CORA}} 
        & {ACC} &26.27 &63.80 &70.43 &50.70 &53.62 & 73.45 &73.88 &\underline{77.07} &\textbf{78.66} \\ 
        &{NMI} &34.68 &47.64 &52.89 &33.78 &46.87 &57.43 &56.45 &\underline{59.21} &\textbf{60.24}\\ 
        &{ARI} &19.35 &38.00 &49.63 &25.76 &30.32 &52.24 &52.51 &\underline{57.52} &\textbf{60.48} \\
        \midrule
        \multirow{3}{*}{{UAT}}
        & {ACC} &42.47 &\underline{56.34} &52.29 &52.25 &51.15 &53.24 &\underline{56.34} &56.04 &\textbf{56.47}\\
        &{NMI} &22.39 &20.69 &21.33 &21.61 &23.47 &26.96 &\textbf{28.15} &26.99 &\underline{27.08}\\
        &{ARI} &15.71 &18.33 &20.50 &21.63 &20.52 &22.49 &\textbf{25.52} &\underline{25.22} & 24.99\\
        \midrule
        \multirow{3}{*}{{AMAP}}
        & {ACC} &36.53 &42.03 &60.14 &71.43 &69.51 &75.25 &\underline{77.25} &77.02 &\textbf{80.16}\\
        & {NMI} &19.31 &31.87 &58.03 &64.13 &60.70 &67.18 &\underline{67.44} &67.21 &\textbf{72.21}\\
        & {ARI} &12.61 &19.31 &43.55 &51.17 &49.09 &56.86 &57.99 &\underline{58.01} &\textbf{63.75}\\
        \midrule
        \multirow{3}{*}{{AMAC}}
        & {ACC} &36.44 &43.14 &49.26 &54.12 &54.92 &\underline{58.43} &53.57 &OOM &\textbf{67.79}\\
        &{NMI} &16.64 &35.47 &39.28 &39.90 &44.36 &\underline{49.92} &34.22 &OOM &\textbf{55.26}\\
        &{ARI} &28.08 &27.06 &35.29 &28.84 &35.61 &\underline{38.29} &32.42 &OOM &\textbf{54.13}\\
        \midrule
        \multirow{3}{*}{{PUBMED}}
        & {ACC} &43.83 &62.09 &68.73 &59.21 &69.51 &\textbf{72.42} &42.58 &OOM &\underline{72.17}\\ 
        & {NMI} &15.05 &23.84 &28.26 &19.65 &31.13 &\underline{35.13} &21.87 &OOM &\textbf{37.02}\\
        & {ARI} &11.43 & 20.62 &29.84 &17.07 &30.85 &\underline{34.19} &21.23 &OOM &\textbf{36.07}\\
        \bottomrule
    \end{tabular}
    \caption{Node clustering performance of nine methods on five datasets. The optimal and sub-optimal results are decorated with bold and underline, respectively. 'OOM' means out-of-memory.}
\label{table3}
\end{table*}
\begin{table}[h]
    \centering
    \small 
    \setlength{\tabcolsep}{2.5pt} 
    \begin{tabular}{l|c|cccccc}
    \toprule
    {Dataset} &{Metric} &{wo/CL} &{wo/CE} &{wo/CUd} &{wo/CUc} &{CCGL}\\  
    \midrule
    \multirow{3}{*}{{CORA}} 
    & {ACC} &77.06 &77.21 &72.60 &71.52 &\textbf{78.66}\\
    & {NMI} &56.43 &57.29 &54.29 &53.34 &\textbf{60.24}\\
    & {ARI} &56.31 &58.05 &51.13 &43.35 &\textbf{60.48}\\
    \midrule
    \multirow{3}{*}{{UAT}}
    & {ACC} &55.46 &55.13 &48.74 &50.92 &\textbf{56.74}\\
    & {NMI} &26.41 &27.09 &25.36 &22.99 &\textbf{27.08}\\
    & {ARI} &24.41 &25.24 &15.17 &19.65 &\textbf{24.99}\\
    \midrule
    \multirow{3}{*}{{AMAP}}
    & {ACC} &77.28 &78.95 &78.17 &77.49 &\textbf{80.16}\\
    & {NMI} &67.28 &71.33 &70.05 &69.60 &\textbf{72.21}\\
    & {ARI} &57.97 &62.43 &59.37 &58.95 &\textbf{63.75}\\
    \midrule
    \multirow{3}{*}{{AMAC}}
    & {ACC} &65.58 &67.37 &59.70 &58.38 &\textbf{67.79}\\
    & {NMI} &55.11 &55.77 &52.58 &55.20 &\textbf{55.26}\\
    & {ARI} &48.06 &51.21 &51.42 &41.00 &\textbf{54.13}\\
    \midrule
    \multirow{3}{*}{{PUBMED}}
    &{ACC} &63.83 &71.91 &61.58 &67.25 &\textbf{72.17}\\
    &{NMI} &29.39 &36.17 &33.68 &30.44 &\textbf{37.02}\\
    & {ARI} &26.06 &35.23 &29.01 &27.94 &\textbf{36.07}\\
    \bottomrule
    \end{tabular}
    \caption{Ablation results of CCGL. The optimal result are shown in bold.}
    \label{table4}
\end{table}

\subsection{Benchmark Datasets}
To substantiate the efficacy of the CCGL, fix publicly accessible real-world datasets are adopted as benchmarks, including CORA, UAT, PUBMED, AMAP, and AMAC. The above datasets are collected from a range of domains such as air traffic, academic citation, and shopping networks. Further details regarding these datasets are shown in Table 1.

\subsection{Evaluation Metrics}
The clustering result is evaluated with three well-known metrics, including Accuracy (ACC), Normalized Mutual Information (NMI), and Average Rand Index (ARI). All metrics are positively correlated with clustering performance, and the range is [0, 100].

\subsection{Comparison with Competitors}
Eight state-of-the-art node clustering algoritms are selected for comparative analysis. This evaluation encompasses a range of approaches, from the traditional K-Means algorithm to advanced GCN-based deep models, such as GAE \cite{kipf2016variational}, DAEGC \cite{wang2019attributed}, and SDCN \cite{bo2020structural}, as well as contrastive learning-based techniques including GCA, SCAGC, CCGC, and HSAN \cite{liu2023hard}.

\subsubsection{Setups.}
The $K$-Means algorithm only utilizes the original node attributes as input. In contrast, other baseline methods use both the original node attributes and the topological graph. The hyper-parameters for each competitor are configured according to the recommendations provided in the original papers. Moreover, to reduce the impact of random factors, each method is executed ten times. For the proposed CCGL, we use the adaptive hyper-parameter selection, which means $\alpha = \frac{\left\| v \right\|_1}{N}$ and $\beta = 1 - \alpha$. Additionally, a parameter grid search is conducted for $\gamma$. The advantage of this approach is that $\alpha$ will increase as the number of nodes involved in the clustering task grows, while $\beta$ for the discrimination task will correspondingly decrease. 

To ensure a fair comparison, each algorithm is executed 10 times to report the average. All deep models are trained with a NVIDIA RTX-4090 GPU.


\begin{figure}[t]
    \centering
    \begin{subfigure}{0.15\textwidth}
        \includegraphics[width=\linewidth]{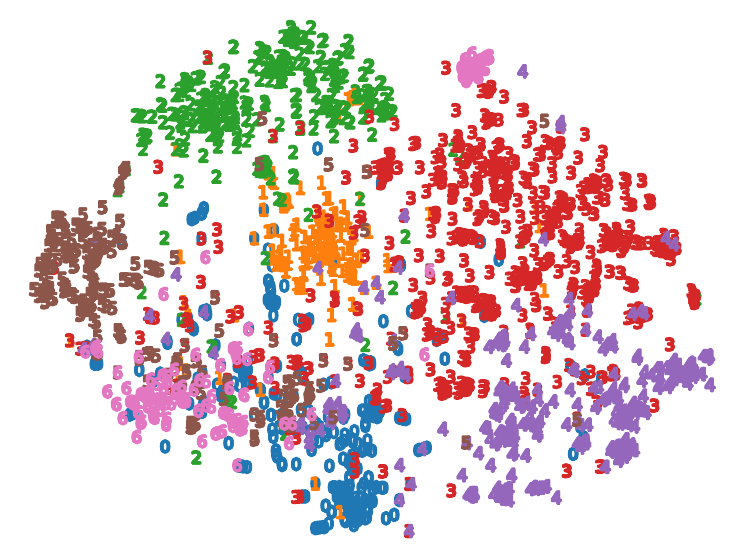}
        \caption{wo/CUc}
        \label{fig:cora}
    \end{subfigure}
    \hspace{0cm} 
    \begin{subfigure}{0.15\textwidth}
        \includegraphics[width=\linewidth]{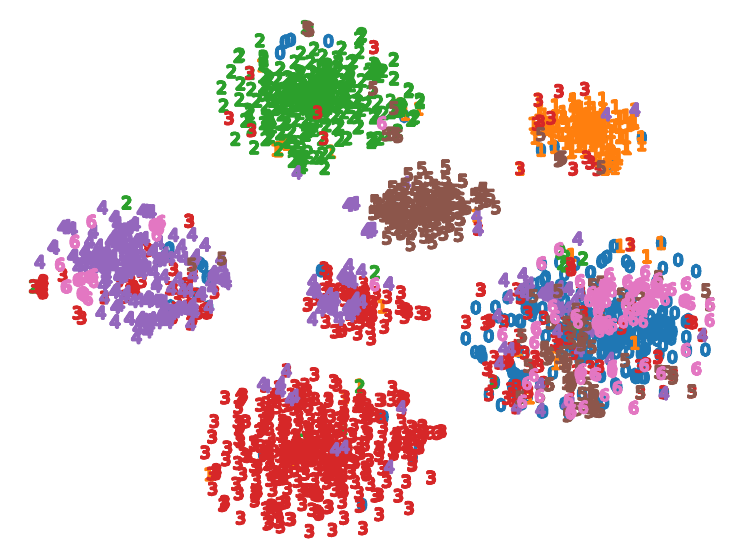}
        \caption{wo/CUd}
        \label{fig:amap}
    \end{subfigure}
    \hspace{0cm} 
    \begin{subfigure}{0.15\textwidth}
        \includegraphics[width=\linewidth]{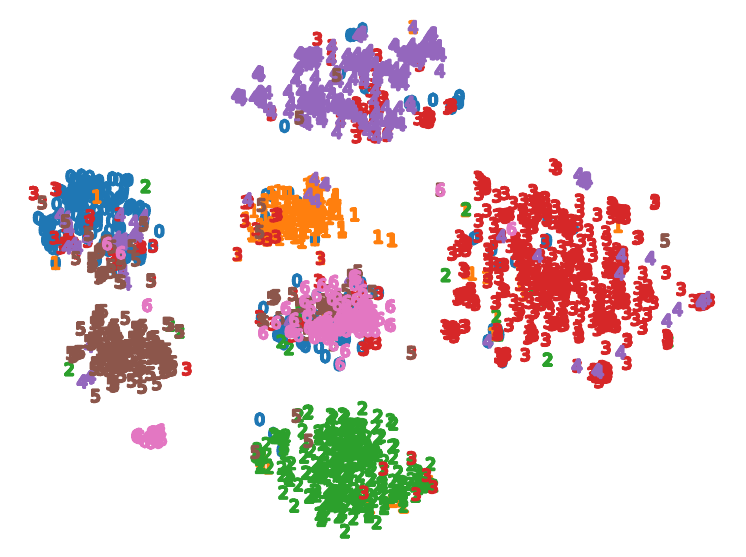}
        \caption{CCGL}
        \label{fig:pubmed}
    \end{subfigure}
    \caption{2D Visualization of learned embeddings on Cora dataset. For better observation, only the first 100 samples of each class are selected.}
    \label{fig:example}
\end{figure}

\subsubsection{Performance Comparison.}

Table 2 displays the average clustering performance of all algorithms. In general, the proposed CCGL outperforms other advanced methods, and presents the best clustering ability on all datasets, which indicates the practicability of CCGL on various graph clustering scenarios. From the experimental results, we also summarize the following viewpoints. Firstly, all GCN-based methods surpass $K$-Means on attributed graph clustering, which manifests the enormous advantage of GCN on graph data mining. GCN-based deep models process the node attributes and the topological structure information simultaneously, so as to detect the internal data clusters more precisely.
Secondly, benefiting from the efficient data augmentation techniques, GCN-based models outperform the traditional graph auto-encoder framework. The augmented view provide a more extensive semantic space for presentation learning, so as to improve clustering. Thirdly, compared to advanced contrastive learning-based baselines, CCGL introduces the clustering guidance to improve the graph augmentation, which makes the augmented views more suitable for the downstream clustering task. Furthermore, unlike existing clustering-oriented methods such as CCGC, SCAGC, and HSAN, the proposed CCGL considers the inherent difficulty distribution of data, and adjusts the contrastive task according to the learning state of the sample by the multi-task curriculum learning scheme. The superior performance of CCGL demonstrates the advantages of curriculum contrastive learning on handling large-scale complex graph data.

\begin{figure}[t]
    \centering
    \begin{subfigure}{0.15\textwidth}
        \includegraphics[width=\linewidth]{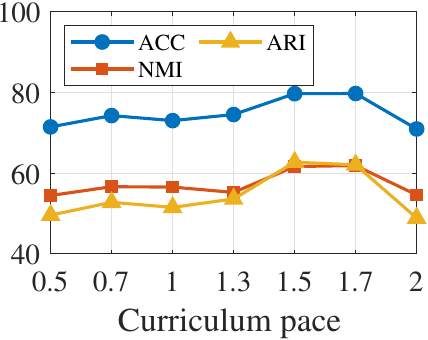}
        \caption{CORA}
        \label{fig:cora}
    \end{subfigure}
    \begin{subfigure}{0.15\textwidth}
        \includegraphics[width=\linewidth]{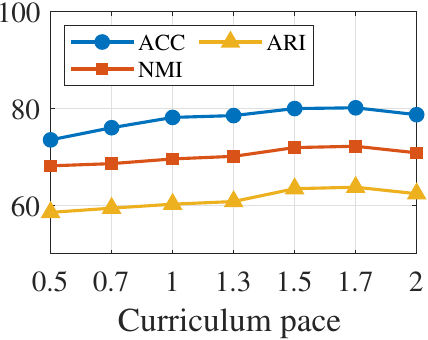}
        \caption{AMAP}
        \label{fig:amap}
    \end{subfigure}
    \begin{subfigure}{0.15\textwidth}
        \includegraphics[width=\linewidth]{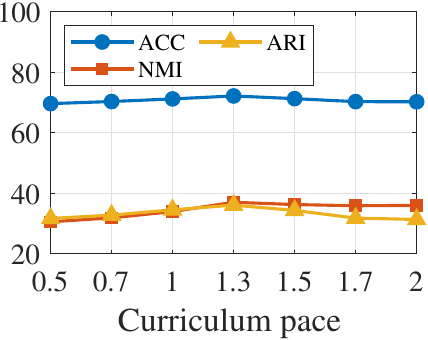}
        \caption{PUBMED}
        \label{fig:pubmed}
    \end{subfigure}
    \caption{Effect of curriculum pace on clustering performance.}
    \label{fig:example}
\end{figure}
\subsection{Ablation Study and Analysis} 
In this part, we conduct the ablation study to verify the effectiveness of the new mechanisms. Three variants of CCGL are designed, including wo/CL, wo/CE, and wo/CUd and wo/CUc. Specifically, in wo/CL, we substitute the cluster-friendly graph augmentation with random data augmentation. In wo/CE, we suspend the clustering entropy loss. Furthermore, in wo/CUd and wo/CUc, we suspend the curriculum learning mechanism, allowing nodes to execute a fixed contrastive learning during model training. wo/CUd represents that all nodes only perform the discrimination task, and wo/CUc represents that all nodes only perform the clustering task. Table 3 presents the ablation comparison on five datasets. It can be seen that CCGL still presents the best clustering scores, which proves the positive effects of the new modules on graph clustering. Furthermore, Fig. 2 shows the visualization results of the embeddings. CCGL presents the best sample distribution with a clear cluster structure. wo/CUc does not present an obvious clustering structure because only discriminative features of samples are learned. Although wo/CUd shows a cluster structure, mixed clusters appear due to the lack of discriminative features of samples.

To further explore the effectiveness of multi-task curriculum learning, we adjust the curriculum pace and fix the task ratio to analyze the performance fluctuation.

\subsubsection{Impact of Curriculum Pace.}
The curriculum pace $\varepsilon$ determines the learning speed. 
When $\varepsilon = 1$, the model iterates to the last epoch, and exactly all nodes are transferred to the clustering task. 
When $\varepsilon < 1 $, not all nodes are transferred to the clustering task until the last epoch, and when $\varepsilon>1$, all nodes are transferred to the clustering task before the end of training. Fig. 3 shows the effect of curriculum pace. The clustering performance reaches the optimal when $1 < \varepsilon < 2$. The range is consistent with the clustering goal of the model, as it tries to perform the clustering task for a period of epochs after all nodes have learned a clear representation with the discrimination task. Too fast curriculum pace also leads to poor clustering, which may be because the sample is quickly transferred to the clustering task while not learning a discriminative embedding adequately beforehand. The experimental result again testifies the feasibility of multi-task contrastive scheme.

\subsubsection{Impact of Task Ratio.}
Finally, we eliminate the curriculum learning mechanism and fix the sample ratio of the two tasks to explore their influence. When the task ratio is 0, all nodes only perform the simple discrimination task during the entire model training. When the task ratio is 1, all nodes engage in the complex clustering task. The results are presented in Fig. 4. For ease of observation, only the comparison in term of ACC and ARI is displayed. Obviously, relying solely on either the discrimination task or clustering task yields sub-optimal performances.
Compared to performing a single task (i.e., task ratio is 0 or 1), multi-task learning with a fixed sample ratio can improve clustering effects, but it is still not as effective as CCGL with adaptive task allocation by curriculum learning. If a node only maintains a fixed contrastive learning, the discrimination task neglects the capture of clustering information, and the clustering task may lead to the accumulation of adverse noise due to the the lack of discriminability in the early stages. To sum up, the combination of discrimination task and clustering task, and adaptive adjustment are beneficial to learning clustering-friendly graph embedding for performance improvement.


\begin{figure}[t]
    \centering
    \begin{subfigure}{0.15\textwidth}
        \includegraphics[width=\linewidth]{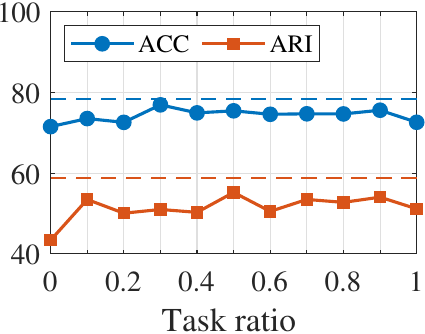}
        \caption{CORA}
        \label{fig:cora}
    \end{subfigure}
    \hspace{0cm} 
    \begin{subfigure}{0.146\textwidth}
        \includegraphics[width=\linewidth]{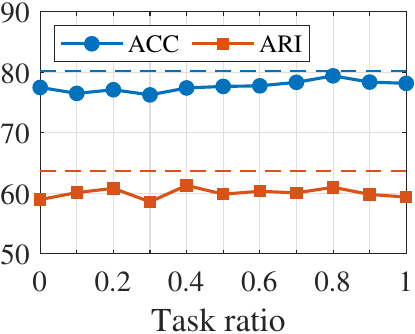}
        \caption{AMAP}
        \label{fig:amap}
    \end{subfigure}
    \hspace{0cm} 
    \begin{subfigure}{0.15\textwidth}
        \includegraphics[width=\linewidth]{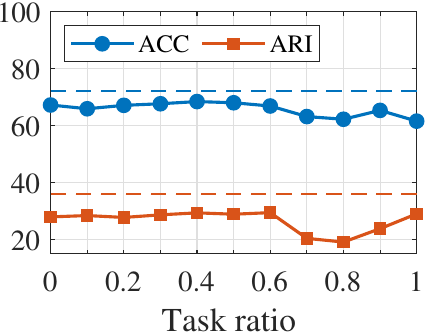}
        \caption{PUBMED}
        \label{fig:pubmed}
    \end{subfigure}
    \caption{Clustering performance of CCGL with a fixed task ratio. The dotted line represents the performance with the automatic task ratio.}
    \label{fig:example}
\end{figure}

\section{Conclusion}

In this paper, we establish a Clustering-guided Curriculum Graph contrastive Learning (CCGL) framework. Clustering entropy is defined based on the embedding clustering results to serve as the clustering guidance for the entire framework. Concretely, we design a clustering-friendly augmentation strategy that reduces the noise caused by the traditional random way, so as to obtain structure-/feature-level graph augmentation results. In addition, the proposed multi-task curriculum learning scheme allows contrastive learning to focus on discrimination task in the early stage, and to pay more attention to  clustering task in the late stage. The flexible scheme adaptively adjusts the selection strategy of positive and negative samples during the training process, which is more suitable for data with complex distribution. The efficiency of our method has been validated by a series of comprehensive experiments.

\bibliography{aaai25}
\section{Reproducibility Checklist}

Unless specified otherwise, please answer “yes” to each question if the relevant information is described either in the paper itself or in a technical appendix with an explicit reference from the main paper. If you wish to explain an answer further, please do so in a section titled “Reproducibility Checklist” at the end of the technical appendix.

\begin{enumerate}
	\item This paper:
	
	\begin{itemize}
		\item Includes a conceptual outline and/or pseudocode description of AI methods introduced (\textbf{yes})
		
		\item Clearly delineates statements that are opinions, hypothesis, and speculation from objective facts and results (\textbf{yes})
		
		\item  Provides well marked pedagogical references for less-familiare readers to gain background necessary to replicate the paper (\textbf{yes})
	\end{itemize}
	
	\item Does this paper make theoretical contributions? (\textbf{no}) \\
	If yes, please complete the list below.
	
	\begin{itemize}
		\item All assumptions and restrictions are stated clearly and formally. 
		
		\item All novel claims are stated formally (e.g., in theorem statements). 
		
		\item Proofs of all novel claims are included. 
		
		\item Proof sketches or intuitions are given for complex and/or novel results. 
		
		\item Appropriate citations to theoretical tools used are given. 
		
		\item All theoretical claims are demonstrated empirically to hold. 
		
		\item All experimental code used to eliminate or disprove claims is included. 
	\end{itemize}
	
	\item Does this paper rely on one or more datasets? (\textbf{yes}) \\
	If yes, please complete the list below.
	
	\begin{itemize}
		\item A motivation is given for why the experiments are conducted on the selected datasets (\textbf{yes})
		
		\item All novel datasets introduced in this paper are included in a data appendix. (\textbf{NA})
		
		\item All novel datasets introduced in this paper will be made publicly available upon publication of the paper with a license that allows free usage for research purposes. (\textbf{NA})
		
		\item All datasets drawn from the existing literature (potentially including authors’ own previously published work) are accompanied by appropriate citations. (\textbf{yes})
		
		\item All datasets drawn from the existing literature (potentially including authors’ own previously published work) are publicly available. (\textbf{yes})
		
		\item All datasets that are not publicly available are described in detail, with explanation why publicly available alternatives are not scientifically satisficing. (\textbf{NA})
	\end{itemize}
	
	\item Does this paper include computational experiments? (\textbf{yes}) \\
	If yes, please complete the list below.
	
	\begin{itemize}
		\item Any code required for pre-processing data is included in the appendix. (\textbf{yes})
		
		\item All source code required for conducting and analyzing the experiments is included in a code appendix. (\textbf{yes})
		
		\item All source code required for conducting and analyzing the experiments will be made publicly available upon publication of the paper with a license that allows free usage for research purposes. (\textbf{yes})
		
		\item All source code implementing new methods have comments detailing the implementation, with references to the paper where each step comes from. (\textbf{yes})
		
		\item If an algorithm depends on randomness, then the method used for setting seeds is described in a way sufficient to allow replication of results. (\textbf{yes})
		
		\item This paper specifies the computing infrastructure used for running experiments (hardware and software), including GPU/CPU models; amount of memory; operating system; names and versions of relevant software libraries and frameworks. (\textbf{yes})
		
		\item This paper formally describes evaluation metrics used and explains the motivation for choosing these metrics. (\textbf{yes})
		
		\item This paper states the number of algorithm runs used to compute each reported result. (\textbf{yes})
		
		\item Analysis of experiments goes beyond single-dimensional summaries of performance (e.g., average; median) to include measures of variation, confidence, or other distributional information. (\textbf{yes})
		
		\item The significance of any improvement or decrease in performance is judged using appropriate statistical tests (e.g., Wilcoxon signed-rank). (\textbf{no})
		
		\item This paper lists all final (hyper-)parameters used for each model/algorithm in the paper’s experiments. (\textbf{yes})
		
		\item This paper states the number and range of values tried per (hyper-) parameter during development of the paper, along with the criterion used for selecting the final parameter setting. (\textbf{yes})
	\end{itemize}
\end{enumerate}
\end{document}